\documentclass[11pt]{article}

\usepackage[preprint]{acl}
\usepackage{colortbl}
\usepackage{times}
\usepackage{latexsym}
\usepackage{booktabs}
\usepackage{graphicx}      
\usepackage{booktabs}      
\usepackage[table]{xcolor} 

\usepackage{fancyhdr}
\usepackage[greek, english]{babel}
\usepackage[T1]{fontenc}

\usepackage[utf8]{inputenc}

\usepackage{microtype}

\usepackage{inconsolata}

\usepackage{graphicx}

\usepackage[most]{tcolorbox} 

\newtcolorbox{promptbox}[1][]{
    colback=gray!5,       
    colframe=gray!60,     
    width=\linewidth,     
    arc=2mm,              
    boxrule=0.5pt,        
    fonttitle=\bfseries,  
    breakable,            
    before=\small,        
    #1                    
}

\title{The Lay of the Land in Idiom-Translation Quality Estimation}

 \author{Malik Marmonier \quad Benoît Sagot \quad Rachel Bawden \\
         Inria, Paris, France\\ \{\texttt{firstname.lastname\}@inria.fr}}

\begin{document}
\maketitle
\begin{abstract}
Idioms have long been a challenge for machine translation, and while large language models (LLMs) have proven increasingly adept at handling them, whether quality estimation (QE) systems have kept pace and can reliably score idiomatic renderings of these expressions above faulty literal ones remains unclear. We introduce the first parallel dataset pairing idiom-bearing English sentences with both literal and idiomatic translations across nine target languages, augmented into multi-sentence segments and a graded literality continuum. Using this dataset, we benchmark eleven conventional QE metrics from the COMET and MetricX families and ten LLMs from the Qwen-3, Gemma-4 and OLMo-3 families used as GEMBA-style judges. We find that most systems, especially older metrics and smaller models, display a marked literality bias, rating literal translations of idioms above idiomatic ones. Higher parameter counts and newer training sets appear to reduce this bias to an extent; nevertheless, of all the benchmarked systems, only MetricX-24-Hybrid (XL, XXL) and Gemma-4-31B consistently rate idiomatic translations of idioms above literal ones, and only Gemma-4-31B reliably ranks intermediate degrees of idiomaticity and literality. Our findings caution against deploying standard QE metrics in domains rich in figurative language, where higher scores might be rewarding undesirable literalism.

\end{abstract}

\section{Introduction}

Idioms are formulaic phrases whose meaning cannot be deduced from the semantic analysis of their isolated constituents: \textit{I have other fish to fry} (`I have more important things to care about'); \textit{they're cutting corners} (`they are trying to save money at the expense of quality and safety'). Idioms can be borrowed through calques between languages in contact, inherited from a common ancestor language, or acquired identically through independent convergence, in which cases they might be translatable literally. Such is rarely the case, however, and idioms generally rule out literal translation, instead requiring the use of complex techniques of paraphrase and cultural equivalence. For this reason, idioms have historically posed a significant challenge in machine translation (MT) research, and were once even cited as proof of the impossibility of MT as a task \citep{bar-hillel-1952-treatment}, with guidelines to academic MT users recommending their removal from system input \citep{bowker-global-2019}.

MT research has since progressed by leaps and bounds \citep{sutskeverSeq2014,bahdanau2015neural,vaswaniAtt2017}, and the shift to large language models (LLMs) has led to substantial improvements in the translation of figurative language and multi-word expressions \citep{raunak-etal-2023-gpts}---categories to which idioms belong---as the large parametric memory of these models and the gargantuan volumes of text data used to train them enhance their ability to learn these constructions. Whether the evaluation and benchmarking apparatus of MT has kept pace remains unclear, however, particularly with respect to quality estimation (QE) techniques. 

QE consists in evaluating the quality of a candidate translation on the sole basis of its source segment, without a gold standard reference translation. Such a task is eminently useful. In real-world applications, the existence of a reference translation would naturally obviate the need for a candidate translation. While the actual extent of QE ``deployment'' in the context of contemporary translation practices is partly obscured by trade secrets, clues within the research output of language service providers and the publications of industry analysts indicate that QE systems are likely assuming a growing role in contemporary translation workflows \citep{rei-etal-2022-cometkiwi, edwards2023automating, guerreiro-etal-2024-xcomet, stasimioti2025does}. In MT research, QE metrics likewise serve an increasing range of functions, from data filtering for in-context learning \citep{pourmostafa-roshan-sharami-etal-2024-guiding} to applications in quality-aware decoding \citep{fernandes-etal-2022-quality,koneru-etal-2025-quality,mrozinski2025qualityestimationrerankingdocumentlevel}. In light of these roles and growing importance, developing a better understanding of the properties and behaviors of dominant QE systems has become an imperative,\footnote{As shown by the recent creation within AMTA of a working group to standardize QE system evaluation \citep{stasimioti2026amta}.} with notable past work analyzing the gender biases these QE systems might display \citep{zaranis-etal-2025-watching}. The present work aims to study the performance of dominant QE approaches in idiom translation.

Through this work, we introduce the first parallel dataset pairing idiom-bearing English sentences with both literal and idiomatic translations across nine target languages, and use it to benchmark eleven conventional QE metrics from the COMET and MetricX families and ten LLMs from the Qwen-3, Gemma-4 and OLMo-3 families used as GEMBA-style judges. We find that most systems, especially older metrics and smaller models, show a marked literality bias, rating literal translations of idioms above idiomatic ones. While larger parameter counts and more recent training sets reduce this bias, of all the benchmarked systems, only MetricX-24-Hybrid (XL, XXL) and Gemma-4-31B consistently rate idiomatic translations of idioms above literal ones, and only Gemma-4-31B reliably ranks intermediate degrees of idiomaticity and literality. Our findings caution against deploying standard QE metrics in domains rich in figurative language, where high scores might be rewarding undesirable literalism.

We publish our dataset under CC BY-SA license.\footnote{The dataset will be available as a password-protected release at \url{https://github.com/mmarmonier/IDIOLIT}.}

\section{Methods}

In this section we describe the methods we applied to the study of QE in idiom translation, presenting the dataset we compiled and the QE metrics we benchmarked using it.

\subsection{Idiom Dataset}
To the best of our knowledge, while there exist many idiom datasets (see Related Work), none of these corpora systematically features paired idiomatic and literal translations of idiom-bearing sentences, in any language. Such a resource would be of precious help to researchers interested in studying a variety of phenomena (e.g., literality bias, literary translation performance), topics of growing importance in MT research \citep{wang-etal-2023-findings, wang-etal-2024-findings, raunak-etal-2023-gpts, kocmi-etal-2025-findings}.

To address this gap, we constructed a parallel dataset across nine target languages: Arabic, Chinese, French, Italian, Japanese, Modern Greek, Russian, Spanish, and Turkish, with English as the source language.

\begin{table*}[t]
\centering
\small
\resizebox{0.75\textwidth}{!}{%
\begin{tabular}{c c p{0.28\linewidth} p{0.28\linewidth} p{0.28\linewidth}}
\toprule
\textbf{ID} & \textbf{Lang} & \textbf{English Source} & \textbf{Literal Translation} & \textbf{Idiomatic Translation} \\
\midrule
1228 & EL & As soon as the bill arrived, he vanished into thin air. & \textgreek{Μόλις έφτασε ο λογαριασμός, εξαφανίστηκε στον λεπτό αέρα.} & \textgreek{Μόλις έφτασε ο λογαριασμός, έγινε καπνός.} \\
\addlinespace
2448 & ES & When he offered to work for free, I started to smell a rat. & \textit{Cuando se ofreció a trabajar gratis, empecé a oler una rata.} & \textit{Cuando se ofreció a trabajar gratis, me empezó a oler a chamusquina.} \\
\addlinespace
2455 & ES & That name doesn't ring a bell. & \textit{Ese nombre no toca una campana.} & \textit{Ese nombre no me suena.} \\
\addlinespace
2437 & ES & I'm so tired, I'm going to hit the sack early tonight. & \textit{Estoy tan cansado, voy a golpear el saco temprano esta noche.} & \textit{Estoy tan cansado, me voy a meter en el sobre temprano esta noche.} \\
\addlinespace
3683 & FR & He will finally pay you back when pigs fly. & \textit{Il te remboursera enfin quand les cochons voleront.} & \textit{Il te remboursera enfin quand les poules auront des dents.} \\
\addlinespace
4889 & IT & She always bends over backwards to help her friends. & \textit{Si piega sempre all'indietro per aiutare i suoi amici.} & \textit{Si fa sempre in quattro per aiutare i suoi amici.} \\
\addlinespace
3677 & FR & He only cleans his apartment once in a blue moon. & \textit{Il ne nettoie son appartement qu'une fois dans une lune bleue.} & \textit{Il ne nettoie son appartement que tous les 36 du mois.} \\
\addlinespace
1220 & EL & I tried reading the instructions, but it's all Greek to me. & \textgreek{Προσπάθησα να διαβάσω τις οδηγίες, αλλά είναι όλα ελληνικά για μένα.} & \textgreek{Προσπάθησα να διαβάσω τις οδηγίες, αλλά μου φαίνονται κινέζικα.} \\
\addlinespace
3688 & FR & He said he was sick? Sick, my foot! & \textit{Il a dit qu'il était malade ? Malade, mon pied !} & \textit{Il a dit qu'il était malade ? Malade, mon œil !} \\
\addlinespace
2458 & ES & He got cold feet right before the wedding. & \textit{Él tuvo pies fríos justo antes de la boda.} & \textit{Él se echó atrás justo antes de la boda.} \\
\addlinespace
\bottomrule
\end{tabular}
}
\caption{Examples of single-sentence base triplets, featuring an English source idiom alongside its highly literal and fluent idiomatic translations.}
\label{tab:base_triplets}
\end{table*}

\begin{table*}[t]
\centering
\small
\resizebox{0.75\textwidth}{!}{%
\begin{tabular}{c c p{0.28\linewidth} p{0.28\linewidth} p{0.28\linewidth}}
\toprule
\textbf{ID} & \textbf{Lang} & \textbf{Concatenated Source ($L=3$)} & \textbf{100\% Literal Target} & \textbf{0\% Literal (Full Idiomatic)} \\
\midrule
2991 & ES & He decided to quit smoking and went cold turkey. Don't worry about the exam, it's a piece of cake. If you help me move, I'll scratch your back later. & \textit{Él decidió dejar de fumar y se volvió pavo frío. No te preocupes por el examen, es un trozo de pastel. Si me ayudas a mudarme, rascaré tu espalda más tarde.} & \textit{Él decidió dejar de fumar y lo dejó de golpe. No te preocupes por el examen, es pan comido. Si me ayudas a mudarme, te devolveré el favor más tarde.} \\
\addlinespace
5243 & IT & Please don't fly off the handle over such a small mistake. I won't be coming to work today, I'm feeling a bit under the weather. Are your ears burning? We were just talking about you. & \textit{Per favore, non volare via dal manico per un errore così piccolo. Oggi non verrò al lavoro, mi sento un po' sotto il tempo. Ti stanno bruciando le orecchie? Stavamo proprio parlando di te.} & \textit{Per favore, non perdere le staffe per un errore così piccolo. Oggi non verrò al lavoro, sono un po' giù di tono. Ti fischiano le orecchie? Stavamo proprio parlando di te.} \\
\addlinespace
1609 & EL & Trying to convince him is like beating a dead horse. When the alarm went off, he was running around like a headless chicken. The manager really put me through the wringer today. & \textgreek{Το να προσπαθείς να τον πείσεις είναι σαν να χτυπάς ένα νεκρό άλογο. Όταν χτύπησε ο συναγερμός, έτρεχε τριγύρω σαν ακέφαλο κοτόπουλο. Ο διευθυντής πραγματικά με πέρασε από τον στίφτη σήμερα.} & \textgreek{Το να προσπαθείς να τον πείσεις είναι πιάσε το αυγό και κούρευ' το. Όταν χτύπησε ο συναγερμός, είχε χάσει τα αυγά και τα πασχάλια. Ο διευθυντής πραγματικά μου έψησε το ψάρι στα χείλη σήμερα.} \\
\addlinespace
1666 & EL & I appreciate that she does not mince her words. I warned him about the danger, but he couldn't care less. He will finally pay you back when pigs fly. & \textgreek{Εκτιμώ ότι δεν ψιλοκόβει τα λόγια της. Τον προειδοποίησα για τον κίνδυνο, αλλά δεν μπορούσε να νοιαστεί λιγότερο. Τελικά θα σε αποπληρώσει όταν τα γουρούνια πετάξουν.} & \textgreek{Εκτιμώ ότι μιλάει έξω από τα δόντια. Τον προειδοποίησα για τον κίνδυνο, αλλά με έγραψε στα παλιά του τα παπούτσια. Για να σε αποπληρώσει, ζήσε Μάη μου να φας τριφύλλι.} \\
\addlinespace
4239 & FR & I have other fish to fry today. He only cleans his apartment once in a blue moon. I worked my fingers to the bone on this project. & \textit{J'ai d'autres poissons à frire, aujourd'hui. Il ne nettoie son appartement qu'une fois dans une lune bleue. J'ai travaillé mes doigts jusqu'à l'os sur ce projet.} & \textit{J'ai d'autres chats à fouetter, aujourd'hui. Il ne nettoie son appartement que tous les 36 du mois. J'ai travaillé d'arrache-pied sur ce projet.} \\
\addlinespace
\bottomrule
\end{tabular}
}
\caption{Examples of augmented multi-sentence segments ($L=3$), demonstrating the combination of base triplets prior to gradient permutation.}
\label{tab:augmented_concatenations}
\end{table*}

\paragraph{Base Triplets} The core of the dataset comprises 163 one-sentence triplets, with each triplet consisting of an English source sentence containing an idiom, paired with both a highly literal translation and a fluent idiomatic translation (see Table~\ref{tab:base_triplets} for a sample). These base triplets were manually curated on the basis of lexicographical resources and the linguistic competence of the authors and their native-speaker colleagues. Each triplet had to meet the two following conditions: 1) the English idiom needed to have a readily identifiable equivalent in the target language and 2) said equivalent target idiom needed to differ from a literal rendering of the English source idiom. As a consequence, the resulting base dataset is not strictly multiparallel since these conditions could not be uniformly met across all target languages for any given English source idiom. The per-target-language distribution of these base triplets is as follows: Spanish (38 triplets), French (20 triplets), and 15 triplets each for the remaining seven languages.

\paragraph{Data Augmentation} We applied a data augmentation strategy to these base triplet instances in order to generate multi-sentence segments. For each target language, we randomly sampled and concatenated base triplets into new instances of varying lengths ($L \in \{2, 3, 4, 5\}$ sentences). We generated exactly $N = 300$ unique combinations for each length, resulting in 1,200 augmented multi-sentence segments per language pair (see Table~\ref{tab:augmented_concatenations} for a data sample).

\paragraph{Literality Gradient} Judgments of literality/idiomaticity exist on a continuous spectrum rather than as a strict dichotomy. To test the models' sensitivity to graded changes in literality, we constructed 300 instances per target language by concatenating three base triplets ($L=3$). For each instance, we generated four target variants in which 0, 1, 2, or all 3 sentences (selected at random) are literal translations and the remainder idiomatic, yielding a literality gradient of 0\%, 33\%, 66\%, and 100\% (see Table~\ref{tab:idiomaticity_gradient} for examples).



\begin{table*}[t]
\centering
\scriptsize
\resizebox{0.75\textwidth}{!}{%
\begin{tabular}{c c p{0.16\linewidth} p{0.16\linewidth} p{0.16\linewidth} p{0.16\linewidth} p{0.16\linewidth}}
\toprule
\textbf{ID} & \textbf{Lang} & \textbf{Source ($L=3$)} & \textbf{100\% Literal} & \textbf{66\% Literal} & \textbf{33\% Literal} & \textbf{0\% Literal (Full Idiomatic)} \\
\midrule
5359 & IT & He never had to work hard; he was born with a silver spoon in his mouth. My brother lives in Australia, so I only see him once in a blue moon. I won't be coming to work today, I'm feeling a bit under the weather. & \textit{Non ha mai dovuto lavorare sodo; è nato con un cucchiaio d'argento in bocca. Mio fratello vive in Australia, quindi lo vedo solo una volta in una luna blu. Oggi non verrò al lavoro, mi sento un po' sotto il tempo.} & \textit{Non ha mai dovuto lavorare sodo; è nato con la camicia. Mio fratello vive in Australia, quindi lo vedo solo una volta in una luna blu. Oggi non verrò al lavoro, mi sento un po' sotto il tempo.} & \textit{Non ha mai dovuto lavorare sodo; è nato con la camicia. Mio fratello vive in Australia, quindi lo vedo solo ogni morte di papa. Oggi non verrò al lavoro, mi sento un po' sotto il tempo.} & \textit{Non ha mai dovuto lavorare sodo; è nato con la camicia. Mio fratello vive in Australia, quindi lo vedo solo ogni morte di papa. Oggi non verrò al lavoro, sono un po' giù di tono.} \\
\addlinespace
1653 & EL & With no security at the door, the concert was a free-for-all. He makes a mountain out of a molehill when he faces a problem. Since she got engaged, she has been on cloud nine. & \textgreek{Χωρίς ασφάλεια στην πόρτα, η συναυλία ήταν ένα ελεύθερο για όλους. Κάνει ένα βουνό από έναν τυφλοπόντικα όταν αντιμετωπίζει ένα πρόβλημα. Από τότε που αρραβωνιάστηκε, είναι στο σύννεφο εννέα.} & \textgreek{Χωρίς ασφάλεια στην πόρτα, η συναυλία ήταν μπάτε σκύλοι αλέστε. Κάνει ένα βουνό από έναν τυφλοπόντικα όταν αντιμετωπίζει ένα πρόβλημα. Από τότε που αρραβωνιάστηκε, είναι στο σύννεφο εννέα.} & \textgreek{Χωρίς ασφάλεια στην πόρτα, η συναυλία ήταν μπάτε σκύλοι αλέστε. Κάνει ένα βουνό από έναν τυφλοπόντικα όταν αντιμετωπίζει ένα πρόβλημα. Από τότε που αρραβωνιάστηκε, πετάει στα σύννεφα.} & \textgreek{Χωρίς ασφάλεια στην πόρτα, η συναυλία ήταν μπάτε σκύλοι αλέστε. Πνίγεται σε μια κουταλιά νερό όταν αντιμετωπίζει ένα πρόβλημα. Από τότε που αρραβωνιάστηκε, πετάει στα σύννεφα.} \\
\addlinespace
4013 & FR & He only cleans his apartment once in a blue moon. Don't worry, they are just creating a storm in a teacup. I know this city like the back of my hand. & \textit{Il ne nettoie son appartement qu'une fois dans une lune bleue. Ne t'inquiète pas, ils créent juste une tempête dans une tasse de thé. Je connais cette ville comme le dos de ma main.} & \textit{Il ne nettoie son appartement qu'une fois dans une lune bleue. Ne t'inquiète pas, ils créent juste une tempête dans une tasse de thé. Je connais cette ville sur le bout des doigts.} & \textit{Il ne nettoie son appartement que tous les 36 du mois. Ne t'inquiète pas, ils créent juste une tempête dans une tasse de thé. Je connais cette ville sur le bout des doigts.} & \textit{Il ne nettoie son appartement que tous les 36 du mois. Ne t'inquiète pas, ils en font tout un fromage. Je connais cette ville sur le bout des doigts.} \\
\addlinespace
5400 & IT & My brother lives in Australia, so I only see him once in a blue moon. She finally let the cat out of the bag and revealed the secret. Buying a new car right now would cost an arm and a leg. & \textit{Mio fratello vive in Australia, quindi lo vedo solo una volta in una luna blu. Alla fine ha lasciato il gatto fuori dalla borsa e ha rivelato il segreto. Comprare una macchina nuova in questo momento costerebbe un braccio e una gamba.} & \textit{Mio fratello vive in Australia, quindi lo vedo solo ogni morte di papa. Alla fine ha lasciato il gatto fuori dalla borsa e ha rivelato il segreto. Comprare una macchina nuova in questo momento costerebbe un braccio e una gamba.} & \textit{Mio fratello vive in Australia, quindi lo vedo solo ogni morte di papa. Alla fine ha vuotato il sacco e ha rivelato il segreto. Comprare una macchina nuova in questo momento costerebbe un braccio e una gamba.} & \textit{Mio fratello vive in Australia, quindi lo vedo solo ogni morte di papa. Alla fine ha vuotato il sacco e ha rivelato il segreto. Comprare una macchina nuova in questo momento costerebbe un occhio della testa.} \\
\addlinespace
2949 & ES & He was so annoying, I was glad to see the back of him. You should invest in different companies; don't put all your eggs in one basket. I believe in being honest and calling a spade a spade. & \textit{Él era tan molesto, estaba contento de verle la espalda. Deberías invertir en diferentes empresas; no pongas todos tus huevos en una canasta. Creo en ser honesto y llamar a una pala una pala.} & \textit{Él era tan molesto, estaba contento de verle la espalda. Deberías invertir en diferentes empresas; no te juegues todo a una sola carta. Creo en ser honesto y llamar a una pala una pala.} & \textit{Él era tan molesto, me alegré de perderlo de vista. Deberías invertir en diferentes empresas; no pongas todos tus huevos en una canasta. Creo en ser honesto y llamar las cosas por su nombre.} & \textit{Él era tan molesto, me alegré de perderlo de vista. Deberías invertir en diferentes empresas; no te juegues todo a una sola carta. Creo en ser honesto y llamar las cosas por su nombre.} \\
\addlinespace

4004 & FR & He ran as fast as he could. Stop beating around the bush and answer the question. She gave up looking for her keys. & \textit{Il a couru aussi vite qu'il le pouvait. Arrête de battre autour du buisson et réponds à la question. Elle a abandonné la recherche de ses clés.} & \textit{Il a couru aussi vite qu'il le pouvait. Arrête de battre autour du buisson et réponds à la question. Elle a renoncé à chercher ses clés.} & \textit{Il a couru à toute vitesse. Arrête de battre autour du buisson et réponds à la question. Elle a renoncé à chercher ses clés.} & \textit{Il a couru à toute vitesse. Arrête de tourner autour du pot et réponds à la question. Elle a renoncé à chercher ses clés.} \\
\addlinespace
\bottomrule
\end{tabular}
}
\caption{Examples of the semantic gradient for concatenated segments ($L=3$). The intermediate steps (66\% and 33\% literal) randomly mix literal and idiomatic sentence-level translations to test granular sensitivity.}
\label{tab:idiomaticity_gradient}
\end{table*}

The compiled dataset comprises a total of 10,963 instances.

\subsection{QE System Query}

To find the ``lay of the land'' in idiom-translation QE, we conducted extensive experiments, prompting a wide range of conventional QE systems and LLMs in a GEMBA-style regime to provide QE scores for the segments in our dataset, before computing deltas between the scores obtained by the literal and idiomatic translations for a given system. The following sections present the QE systems we queried.

\subsubsection{Conventional QE Systems}
Conventional QE relies on dedicated models trained to regress on human judgments from past WMT evaluation campaigns, chiefly direct assessment (DA) scores \citep{graham-etal-2013-continuous} and the more fine-grained MQM annotations \citep{burchardt-2013-multidimensional}. We tested eleven such systems from two dominant families. From the COMET family, we used \textbf{wmt22-cometkiwi-da}\footnote{Unbabel/wmt22-cometkiwi-da} \citep{rei-etal-2022-cometkiwi}, built on top of InfoXLM \citep[approx. 600M parameters; ][]{chi-etal-2021-infoxlm}; its XL and XXL successors \textbf{wmt23-cometkiwi-da-xl}\footnote{Unbabel/wmt23-cometkiwi-da-xl} and \textbf{wmt23-cometkiwi-da-xxl}\footnote{Unbabel/wmt23-cometkiwi-da-xxl} \citep{rei-etal-2023-scaling}, built on XLM-R XL and XXL \citep[3.5B and 10.5B parameters, respectively; ][]{conneau-etal-2020-unsupervised}; and \textbf{XCOMET-XL}\footnote{Unbabel/XCOMET-XL} and \textbf{XCOMET-XXL}\footnote{Unbabel/XCOMET-XXL} \citep{guerreiro-etal-2024-xcomet}, which unify reference-based and reference-free evaluation with error span detection while relying on the same base models. Ensembles of these models ranked first at the WMT 2022–2024 QE shared tasks \citep{zerva-etal-2022-findings,blain-etal-2023-findings,zerva-etal-2024-findings}. From the MetricX family, built on top of mT5 \citep{xue-etal-2021-mt5}, we used \textbf{metricx-23-qe-large},\footnote{google/metricx-23-qe-large-v2p0} \textbf{metricx-23-qe-xl}\footnote{google/metricx-23-qe-xl-v2p0} and \textbf{metricx-23-qe-xxl}\footnote{google/metricx-23-qe-xxl-v2p0} \citep[1.2B, 3.7B and 13B parameters, respectively; ][]{juraska-etal-2023-metricx}, and \textbf{metricx-24-hybrid-large},\footnote{google/metricx-24-hybrid-large-v2p6} \textbf{metricx-24-hybrid-xl},\footnote{google/metricx-24-hybrid-xl-v2p6} \textbf{metricx-24-hybrid-xxl}\footnote{google/metricx-24-hybrid-xxl-v2p6-bfloat16} \citep[1.2B, 3.7B and 13B parameters, respectively; ][]{juraska-etal-2024-metricx}, the latter additionally trained on synthetic data covering failure modes underrepresented in WMT data (e.g., fluent but unrelated translations).\footnote{MetricX-25 \citep{juraska-etal-2025-metricx} was not open-sourced and could not be included.} COMET models were run with the \texttt{unbabel-comet} library,\footnote{\href{https://github.com/Unbabel/COMET}{https://github.com/Unbabel/COMET}} and MetricX models through the Transformers library \citep{wolf-etal-2020-transformers}. All models were run at full precision, except metricx-24-hybrid-xxl, for which we used the official bfloat16 checkpoint due to hardware constraints.

\subsubsection{LLMs}
LLM-as-a-judge approaches have achieved increasingly competitive results in dedicated QE shared tasks \citep{zerva-etal-2024-findings, lavie-etal-2025-findings}, notably through GEMBA-style prompting \citep{kocmi-federmann-2023-gemba}. Following \citet{zaranis-etal-2025-watching}, we adopted zero-shot DA-style prompts for our task, presenting the LLM with an English source segment containing an idiom and a candidate translation (either literal or idiomatic) from our dataset, demanding that it output a score between 0 (``no meaning preserved'') and 100 (``perfect meaning and grammar''). The full prompt templates are given in Appendix~\ref{app:prompt}. 

We selected a total of 10 LLMs as QE judges, from 3 model families: \texttt{\textbf{Qwen-3-32B}} \cite{yang2025qwen3technicalreport}, a reasoning LLM with 32 billion parameters, and \texttt{\textbf{Qwen-3-14B}}, \texttt{\textbf{Qwen-3-8B}}, \texttt{\textbf{Qwen-3-1.7B}}, and \texttt{\textbf{Qwen-3-0.6B}}, its smaller distilled variants; \texttt{\textbf{Olmo-3-32B-Think}} \cite{olmo2025olmo3}, a reasoning LLM with 32 billion parameters, and \texttt{\textbf{Olmo-3-7B-Think}}, its smaller variant; \texttt{\textbf{Gemma-4-31B-it}} \cite{gemma4modelcard2026}, a reasoning LLM with 31 billion parameters, and its smaller edge-optimized variants, \texttt{\textbf{Gemma-4-E4B-it}} and \texttt{\textbf{Gemma-4-E2B-it}}. 

To investigate the impact of reasoning on the task at hand, the Qwen-3 and Gemma-4 models were prompted under two distinct settings, one with their reasoning mode turned on and the other with this mode turned off, using slightly different prompt templates and hyperparameters (see Appendix~\ref{app:prompt}). Inference was conducted using H100 GPU nodes, the vLLM framework \citep{kwon2023efficient} and the Transformers library \citep{wolf-etal-2020-transformers} with greedy decoding ($t=0$), and the \texttt{max-tokens} parameter set to 50 for the reasoning-off setting, and to 16,384 tokens for the reasoning-on setting.

\section{Results}

\begin{table*}[t]
\centering
\scriptsize
\resizebox{0.85\textwidth}{!}{%
\begin{tabular}{l c c c c c c c c c }
\toprule
\textbf{QE System} & \textbf{en--ar} & \textbf{en--el} & \textbf{en--es} & \textbf{en--fr} & \textbf{en--it} & \textbf{en--ja} & \textbf{en--ru} & \textbf{en--tr} & \textbf{en--zh} \\
\midrule
wmt22\_cometkiwi\_qe & \cellcolor[RGB]{179,3,38} \color{white} 1.230 & \cellcolor[RGB]{179,3,38} \color{white} 1.449 & \cellcolor[RGB]{235,127,99}  1.093 & \cellcolor[RGB]{179,3,38} \color{white} 1.221 & \cellcolor[RGB]{179,3,38} \color{white} 1.424 & \cellcolor[RGB]{240,141,111}  1.084 & \cellcolor[RGB]{179,3,38} \color{white} 1.232 & \cellcolor[RGB]{239,137,108}  1.087 & \cellcolor[RGB]{246,166,135}  1.065 \\
\midrule
wmt23\_cometkiwi\_xl & \cellcolor[RGB]{179,3,38} \color{white} 1.237 & \cellcolor[RGB]{179,3,38} \color{white} 2.338 & \cellcolor[RGB]{229,216,208}  1.010 & \cellcolor[RGB]{179,3,38} \color{white} 1.784 & \cellcolor[RGB]{179,3,38} \color{white} 1.959 & \cellcolor[RGB]{179,3,38} \color{white} 1.263 & \cellcolor[RGB]{179,3,38} \color{white} 1.296 & \cellcolor[RGB]{179,3,38} \color{white} 1.224 & \cellcolor[RGB]{96,128,232} \color{white} 0.887 \\
wmt23\_cometkiwi\_xxl & \cellcolor[RGB]{179,3,38} \color{white} 1.183 & \cellcolor[RGB]{179,3,38} \color{white} 1.949 & \cellcolor[RGB]{58,76,192} \color{white} 0.647 & \cellcolor[RGB]{179,3,38} \color{white} 1.248 & \cellcolor[RGB]{179,3,38} \color{white} 1.297 & \cellcolor[RGB]{211,219,230}  0.990 & \cellcolor[RGB]{58,76,192} \color{white} 0.754 & \cellcolor[RGB]{58,76,192} \color{white} 0.682 & \cellcolor[RGB]{58,76,192} \color{white} 0.479 \\
\midrule
xcomet\_xl & \cellcolor[RGB]{179,3,38} \color{white} 1.499 & \cellcolor[RGB]{179,3,38} \color{white} 2.301 & \cellcolor[RGB]{244,195,171}  1.038 & \cellcolor[RGB]{179,3,38} \color{white} 1.498 & \cellcolor[RGB]{179,3,38} \color{white} 1.922 & \cellcolor[RGB]{179,3,38} \color{white} 1.289 & \cellcolor[RGB]{179,3,38} \color{white} 1.338 & \cellcolor[RGB]{179,3,38} \color{white} 1.288 & \cellcolor[RGB]{119,154,246}  0.906 \\
xcomet\_xxl & \cellcolor[RGB]{107,141,240} \color{white} 0.896 & \cellcolor[RGB]{207,70,61} \color{white} 1.125 & \cellcolor[RGB]{58,76,192} \color{white} 0.778 & \cellcolor[RGB]{193,212,244}  0.971 & \cellcolor[RGB]{241,202,182}  1.030 & \cellcolor[RGB]{128,164,250}  0.916 & \cellcolor[RGB]{58,76,192} \color{white} 0.788 & \cellcolor[RGB]{58,76,192} \color{white} 0.685 & \cellcolor[RGB]{58,76,192} \color{white} 0.675 \\
\midrule
metricx\_23\_qe\_large & \cellcolor[RGB]{179,3,38} \color{white} 1.587 & \cellcolor[RGB]{179,3,38} \color{white} 1.921 & \cellcolor[RGB]{179,3,38} \color{white} 1.443 & \cellcolor[RGB]{179,3,38} \color{white} 1.607 & \cellcolor[RGB]{179,3,38} \color{white} 2.287 & \cellcolor[RGB]{179,3,38} \color{white} 1.682 & \cellcolor[RGB]{179,3,38} \color{white} 1.500 & \cellcolor[RGB]{179,3,38} \color{white} 1.299 & \cellcolor[RGB]{179,3,38} \color{white} 1.775 \\
metricx\_23\_qe\_xl & \cellcolor[RGB]{179,3,38} \color{white} 1.477 & \cellcolor[RGB]{179,3,38} \color{white} 1.917 & \cellcolor[RGB]{179,3,38} \color{white} 1.390 & \cellcolor[RGB]{179,3,38} \color{white} 1.694 & \cellcolor[RGB]{179,3,38} \color{white} 1.829 & \cellcolor[RGB]{179,3,38} \color{white} 1.509 & \cellcolor[RGB]{179,3,38} \color{white} 1.406 & \cellcolor[RGB]{238,135,106}  1.088 & \cellcolor[RGB]{179,3,38} \color{white} 1.193 \\
metricx\_23\_qe\_xxl & \cellcolor[RGB]{242,200,179}  1.033 & \cellcolor[RGB]{179,3,38} \color{white} 1.316 & \cellcolor[RGB]{179,3,38} \color{white} 1.206 & \cellcolor[RGB]{179,3,38} \color{white} 1.403 & \cellcolor[RGB]{179,3,38} \color{white} 1.724 & \cellcolor[RGB]{179,3,38} \color{white} 1.557 & \cellcolor[RGB]{179,3,38} \color{white} 1.152 & \cellcolor[RGB]{242,199,178}  1.033 & \cellcolor[RGB]{85,113,222} \color{white} 0.877 \\
\midrule
metricx\_24\_hybrid\_large & \cellcolor[RGB]{245,160,129}  1.069 & \cellcolor[RGB]{179,3,38} \color{white} 1.263 & \cellcolor[RGB]{244,196,173}  1.037 & \cellcolor[RGB]{179,3,38} \color{white} 1.202 & \cellcolor[RGB]{179,3,38} \color{white} 1.512 & \cellcolor[RGB]{243,149,118}  1.078 & \cellcolor[RGB]{204,216,237}  0.982 & \cellcolor[RGB]{229,216,208}  1.010 & \cellcolor[RGB]{202,216,238}  0.981 \\
metricx\_24\_hybrid\_xl & \cellcolor[RGB]{146,180,254}  0.930 & \cellcolor[RGB]{58,76,192} \color{white} 0.847 & \cellcolor[RGB]{58,76,192} \color{white} 0.834 & \cellcolor[RGB]{199,214,240}  0.977 & \cellcolor[RGB]{188,209,246}  0.966 & \cellcolor[RGB]{151,184,254}  0.934 & \cellcolor[RGB]{58,76,192} \color{white} 0.826 & \cellcolor[RGB]{58,76,192} \color{white} 0.841 & \cellcolor[RGB]{58,76,192} \color{white} 0.654 \\
metricx\_24\_hybrid\_xxl & \cellcolor[RGB]{58,76,192} \color{white} 0.635 & \cellcolor[RGB]{58,76,192} \color{white} 0.509 & \cellcolor[RGB]{58,76,192} \color{white} 0.496 & \cellcolor[RGB]{58,76,192} \color{white} 0.759 & \cellcolor[RGB]{58,76,192} \color{white} 0.767 & \cellcolor[RGB]{58,76,192} \color{white} 0.734 & \cellcolor[RGB]{58,76,192} \color{white} 0.424 & \cellcolor[RGB]{58,76,192} \color{white} 0.531 & \cellcolor[RGB]{58,76,192} \color{white} 0.374 \\
\midrule
gemba\_qwen3\_0.6b\_noThink & \cellcolor[RGB]{221,220,219}  1.000 & \cellcolor[RGB]{222,219,218}  1.002 & \cellcolor[RGB]{221,220,219}  1.000 & \cellcolor[RGB]{220,220,221}  1.000 & \cellcolor[RGB]{220,220,221}  1.000 & \cellcolor[RGB]{231,214,204}  1.013 & \cellcolor[RGB]{221,220,219}  1.000 & \cellcolor[RGB]{222,219,218}  1.002 & \cellcolor[RGB]{226,217,212}  1.006 \\
gemba\_qwen3\_0.6b & \cellcolor[RGB]{235,211,198}  1.018 & \cellcolor[RGB]{242,201,181}  1.031 & \cellcolor[RGB]{231,214,205}  1.012 & \cellcolor[RGB]{236,210,196}  1.019 & \cellcolor[RGB]{245,161,130}  1.068 & \cellcolor[RGB]{245,193,168}  1.041 & \cellcolor[RGB]{240,204,185}  1.027 & \cellcolor[RGB]{228,216,209}  1.008 & \cellcolor[RGB]{215,219,226}  0.994 \\
gemba\_qwen3\_1.7b\_noThink & \cellcolor[RGB]{229,112,87} \color{white} 1.102 & \cellcolor[RGB]{179,3,38} \color{white} 1.384 & \cellcolor[RGB]{237,208,193}  1.022 & \cellcolor[RGB]{242,145,115}  1.081 & \cellcolor[RGB]{205,66,58} \color{white} 1.129 & \cellcolor[RGB]{237,208,193}  1.022 & \cellcolor[RGB]{237,208,193}  1.022 & \cellcolor[RGB]{246,164,134}  1.066 & \cellcolor[RGB]{174,201,252}  0.954 \\
gemba\_qwen3\_1.7b & \cellcolor[RGB]{179,3,38} \color{white} 1.228 & \cellcolor[RGB]{179,3,38} \color{white} 1.793 & \cellcolor[RGB]{244,195,171}  1.038 & \cellcolor[RGB]{233,121,94}  1.097 & \cellcolor[RGB]{228,110,86} \color{white} 1.103 & \cellcolor[RGB]{247,177,148}  1.055 & \cellcolor[RGB]{241,203,184}  1.028 & \cellcolor[RGB]{179,3,38} \color{white} 1.168 & \cellcolor[RGB]{101,134,236} \color{white} 0.891 \\
gemba\_qwen3\_8b\_noThink & \cellcolor[RGB]{243,149,118}  1.078 & \cellcolor[RGB]{179,3,38} \color{white} 1.936 & \cellcolor[RGB]{166,195,253}  0.947 & \cellcolor[RGB]{153,186,254}  0.937 & \cellcolor[RGB]{137,172,252}  0.922 & \cellcolor[RGB]{130,165,251}  0.916 & \cellcolor[RGB]{113,148,244}  0.902 & \cellcolor[RGB]{99,131,234} \color{white} 0.889 & \cellcolor[RGB]{58,76,192} \color{white} 0.817 \\
gemba\_qwen3\_8b & \cellcolor[RGB]{179,3,38} \color{white} 1.203 & \cellcolor[RGB]{179,3,38} \color{white} 2.300 & \cellcolor[RGB]{124,160,249}  0.912 & \cellcolor[RGB]{167,196,253}  0.948 & \cellcolor[RGB]{126,161,249}  0.913 & \cellcolor[RGB]{119,154,246}  0.907 & \cellcolor[RGB]{115,149,244}  0.903 & \cellcolor[RGB]{233,212,201}  1.015 & \cellcolor[RGB]{58,76,192} \color{white} 0.812 \\
gemba\_qwen3\_14b\_noThink & \cellcolor[RGB]{229,216,208}  1.010 & \cellcolor[RGB]{230,116,90} \color{white} 1.100 & \cellcolor[RGB]{188,209,246}  0.967 & \cellcolor[RGB]{176,203,251}  0.956 & \cellcolor[RGB]{208,218,233}  0.987 & \cellcolor[RGB]{122,157,248}  0.909 & \cellcolor[RGB]{195,213,242}  0.974 & \cellcolor[RGB]{178,203,251}  0.957 & \cellcolor[RGB]{58,76,192} \color{white} 0.777 \\
gemba\_qwen3\_14b & \cellcolor[RGB]{241,144,114}  1.082 & \cellcolor[RGB]{179,3,38} \color{white} 1.357 & \cellcolor[RGB]{149,183,254}  0.932 & \cellcolor[RGB]{175,202,251}  0.954 & \cellcolor[RGB]{155,187,254}  0.938 & \cellcolor[RGB]{63,83,198} \color{white} 0.856 & \cellcolor[RGB]{151,184,254}  0.934 & \cellcolor[RGB]{218,220,223}  0.997 & \cellcolor[RGB]{58,76,192} \color{white} 0.775 \\
gemba\_qwen3\_32b\_noThink & \cellcolor[RGB]{237,208,193}  1.022 & \cellcolor[RGB]{236,130,102}  1.091 & \cellcolor[RGB]{153,186,254}  0.936 & \cellcolor[RGB]{171,199,252}  0.951 & \cellcolor[RGB]{201,215,238}  0.980 & \cellcolor[RGB]{119,154,246}  0.907 & \cellcolor[RGB]{185,208,248}  0.964 & \cellcolor[RGB]{218,220,223}  0.997 & \cellcolor[RGB]{58,76,192} \color{white} 0.743 \\
gemba\_qwen3\_32b & \cellcolor[RGB]{244,195,171}  1.038 & \cellcolor[RGB]{179,3,38} \color{white} 1.180 & \cellcolor[RGB]{108,142,241} \color{white} 0.898 & \cellcolor[RGB]{130,165,251}  0.916 & \cellcolor[RGB]{155,187,254}  0.937 & \cellcolor[RGB]{58,76,192} \color{white} 0.839 & \cellcolor[RGB]{142,177,253}  0.927 & \cellcolor[RGB]{229,216,208}  1.010 & \cellcolor[RGB]{58,76,192} \color{white} 0.719 \\
\midrule
gemba\_gemma4\_e2b\_noThink & \cellcolor[RGB]{200,215,239}  0.978 & \cellcolor[RGB]{179,3,38} \color{white} 2.117 & \cellcolor[RGB]{60,79,195} \color{white} 0.853 & \cellcolor[RGB]{77,103,215} \color{white} 0.870 & \cellcolor[RGB]{234,211,199}  1.017 & \cellcolor[RGB]{58,76,192} \color{white} 0.709 & \cellcolor[RGB]{88,118,226} \color{white} 0.879 & \cellcolor[RGB]{67,90,204} \color{white} 0.860 & \cellcolor[RGB]{58,76,192} \color{white} 0.659 \\
gemba\_gemma4\_e2b & \cellcolor[RGB]{179,3,38} \color{white} 1.295 & \cellcolor[RGB]{179,3,38} \color{white} 3.238 & \cellcolor[RGB]{58,76,192} \color{white} 0.810 & \cellcolor[RGB]{60,79,195} \color{white} 0.853 & \cellcolor[RGB]{185,22,42} \color{white} 1.144 & \cellcolor[RGB]{58,76,192} \color{white} 0.741 & \cellcolor[RGB]{58,76,192} \color{white} 0.787 & \cellcolor[RGB]{99,131,234} \color{white} 0.890 & \cellcolor[RGB]{58,76,192} \color{white} 0.623 \\
gemba\_gemma4\_e4b\_noThink & \cellcolor[RGB]{133,168,251}  0.919 & \cellcolor[RGB]{179,3,38} \color{white} 1.438 & \cellcolor[RGB]{58,76,192} \color{white} 0.773 & \cellcolor[RGB]{58,76,192} \color{white} 0.807 & \cellcolor[RGB]{81,108,219} \color{white} 0.873 & \cellcolor[RGB]{58,76,192} \color{white} 0.689 & \cellcolor[RGB]{58,76,192} \color{white} 0.727 & \cellcolor[RGB]{58,76,192} \color{white} 0.793 & \cellcolor[RGB]{58,76,192} \color{white} 0.591 \\
gemba\_gemma4\_e4b & \cellcolor[RGB]{156,188,254}  0.938 & \cellcolor[RGB]{179,3,38} \color{white} 1.737 & \cellcolor[RGB]{58,76,192} \color{white} 0.707 & \cellcolor[RGB]{58,76,192} \color{white} 0.788 & \cellcolor[RGB]{58,76,192} \color{white} 0.825 & \cellcolor[RGB]{58,76,192} \color{white} 0.676 & \cellcolor[RGB]{58,76,192} \color{white} 0.703 & \cellcolor[RGB]{58,76,192} \color{white} 0.836 & \cellcolor[RGB]{58,76,192} \color{white} 0.577 \\
gemba\_gemma4\_31b\_noThink & \cellcolor[RGB]{58,76,192} \color{white} 0.472 & \cellcolor[RGB]{58,76,192} \color{white} 0.578 & \cellcolor[RGB]{58,76,192} \color{white} 0.489 & \cellcolor[RGB]{58,76,192} \color{white} 0.547 & \cellcolor[RGB]{58,76,192} \color{white} 0.456 & \cellcolor[RGB]{58,76,192} \color{white} 0.317 & \cellcolor[RGB]{58,76,192} \color{white} 0.442 & \cellcolor[RGB]{58,76,192} \color{white} 0.297 & \cellcolor[RGB]{58,76,192} \color{white} 0.342 \\
gemba\_gemma4\_31b & \cellcolor[RGB]{58,76,192} \color{white} 0.413 & \cellcolor[RGB]{58,76,192} \color{white} 0.493 & \cellcolor[RGB]{58,76,192} \color{white} 0.407 & \cellcolor[RGB]{58,76,192} \color{white} 0.465 & \cellcolor[RGB]{58,76,192} \color{white} 0.429 & \cellcolor[RGB]{58,76,192} \color{white} 0.277 & \cellcolor[RGB]{58,76,192} \color{white} 0.375 & \cellcolor[RGB]{58,76,192} \color{white} 0.277 & \cellcolor[RGB]{58,76,192} \color{white} 0.281 \\
\midrule
gemba\_olmo3\_7b & \cellcolor[RGB]{179,3,38} \color{white} 1.437 & \cellcolor[RGB]{179,3,38} \color{white} 1.991 & \cellcolor[RGB]{179,3,38} \color{white} 1.151 & \cellcolor[RGB]{179,3,38} \color{white} 1.184 & \cellcolor[RGB]{179,3,38} \color{white} 1.494 & \cellcolor[RGB]{179,3,38} \color{white} 1.272 & \cellcolor[RGB]{179,3,38} \color{white} 1.166 & \cellcolor[RGB]{179,3,38} \color{white} 1.236 & \cellcolor[RGB]{172,200,252}  0.953 \\
gemba\_olmo3\_32b & \cellcolor[RGB]{179,3,38} \color{white} 1.319 & \cellcolor[RGB]{179,3,38} \color{white} 1.603 & \cellcolor[RGB]{207,217,234}  0.985 & \cellcolor[RGB]{244,157,126}  1.072 & \cellcolor[RGB]{179,3,38} \color{white} 1.300 & \cellcolor[RGB]{198,53,52} \color{white} 1.134 & \cellcolor[RGB]{241,202,182}  1.030 & \cellcolor[RGB]{231,214,205}  1.013 & \cellcolor[RGB]{58,76,192} \color{white} 0.727 \\
\bottomrule
\end{tabular}%
}
\caption{Macro-preference ratio of QE models (Literal / Idiomatic). Values $>1$ indicate a literal bias, while values $<1$ indicate an idiomatic preference. The color scale saturates at 0.85 (blue) and 1.15 (red), as the direction of the model's preference is more informative than the magnitude of the score difference in binary judgments. }
\label{tab:macro_ratio}
\end{table*}

In order to study the QE systems' behaviors in idiom translation, we calculated the per-language-pair corpus-level preference ratio for each evaluated system. For quality metrics (all COMET-derived systems and GEMBA scores), this is expressed as the ratio of the mean score assigned to fully literal translations over the mean score assigned to the fully idiomatic translations. For error-based metrics (MetricX), the ratio is inverted for legibility purposes: 
\begin{equation}
\resizebox{\columnwidth}{!}{$
    \mathrm{PR} = 
    \begin{cases}
        \dfrac{\bar{s}_{\mathrm{lit}}}{\bar{s}_{\mathrm{idio}}} & \text{for quality-based metrics (COMET, GEMBA)}, \\[10pt]
        \dfrac{\bar{s}_{\mathrm{idio}}}{\bar{s}_{\mathrm{lit}}} & \text{for error-based metrics (MetricX)},
    \end{cases}
$}
\label{eq:preference_ratio}
\end{equation}
where $\bar{s}_{\mathrm{lit}}$ and $\bar{s}_{\mathrm{idio}}$ denote the mean scores assigned to fully literal and fully idiomatic translations across the corpus. Consequently, a ratio greater than $1.0$ indicates a literality bias whereby the QE system rates the literal translation higher than the idiomatic one, a ratio less than $1.0$ indicates an idiomatic preference, and a ratio of exactly $1.0$ indicates neutral scoring.

Table~\ref{tab:macro_ratio} displays these ratios, from which we draw the following observations.\footnote{Appendix~\ref{app:bootstrap} also presents a system-wide comparison of preference ratio deltas with significance tests, obtained through bootstrap resampling \citep{koehn-2004-statistical}.}

An analysis of the ``conventional'' COMET and MetricX QE systems reveals that the ability to reward idiomaticity in idiom translation---and, conversely, to penalize literality---is driven by an interplay between model scale and finetuning data composition. The impact of the latter is evidenced by the observable ``generational'' shift within the MetricX system family: despite matching parameter counts across their Large (1.2B), XL (3.7B), and XXL (13B) mT5-derived checkpoints, the MetricX-24-Hybrid group effectively suppresses the severe literality bias exhibited by its predecessor, MetricX-23-QE. This improvement is likely attributable to MetricX-24's exposure to larger MQM training sets and/or to the hybrid nature of the model, capable of performing both QE and reference-based evaluation. A similar though fainter trend of improvement is visible between the WMT23 COMET-Kiwi models and the newer XCOMET architecture. Concurrently, a scaling law appears to govern QE performance in idiom translation. Scaling up parameter counts---such as moving from Large to XXL in MetricX-24, or from XL (3.5B) to XXL (10.5B) in XCOMET---consistently shifts the metrics away from a literal bias and toward an idiomatic preference. This is likely due to the non-compositional nature of idioms: since their meaning cannot be deduced analytically, recognizing idioms and their idiomatic equivalent in a target language depends on memorized knowledge of specific expressions, which inherently benefits from the increased parametric memory of larger models. Ultimately, reliable idiomatic sensitivity in conventional QE systems is currently only achieved by the XL and XXL checkpoint of MetricX-24-Hybrid.

When using LLMs as QE judges under the GEMBA framework, sensitivity to idiomaticity is characterized by stark inter-family variance and diverse scaling trajectories. The Qwen 3 family exhibits a distinct, non-linear progression: its smallest checkpoint (0.6B) gives preference ratios hovering tightly around 1.0, indicating that it is effectively blind to the distinction between literal and idiomatic phrasing. As parameter count increases (1.7B and 8B), the models develop a severe literality bias, before eventually tilting toward an idiomatic preference at the 32B scale for a majority of the target languages, though not for their totality, with Arabic and Greek eliciting a literality bias and Turkish neutral judgments. Conversely, the Gemma 4 family bypasses this initial blindness entirely. Even its smallest edge-optimized model (E2B) actively differentiates translations, albeit unreliably---exhibiting an extreme literality bias for target segments in Greek (incidentally the lowest-resource target language represented in our dataset) and to a lesser extent in Arabic and Italian with its reasoning mode switched on. Scaling up the Gemma parameter count rapidly stabilizes this behavior: Gemma-4-E4B achieves broad idiomatic preference, albeit with a persistent literality bias in Greek, and Gemma-4-31B demonstrates a uniform robust preference for idiomaticity across all target languages. Parameter scale does not compensate for everything, however. Although OLMo-3-32B shows a weaker literality bias than its 7B counterpart, it remains well behind Qwen-3 and Gemma-4 models of comparable size. OLMo differs from these models in its training data, both in pretraining and post-training, and in particular in its far more limited multilingual coverage. This suggests that scale helps only insofar as the relevant idiomatic knowledge has been acquired during training.

LLMs as QE judges therefore have checkered performance, with Gemma-4-31B as the only LLM reliably capable of penalizing literality in idiom translation over idiomaticity.

\begin{table*}[t]
\centering
\scriptsize
\resizebox{0.85\textwidth}{!}{%
\begin{tabular}{l c c c c c c c c c }
\toprule
\textbf{QE System} & \textbf{ar} & \textbf{el} & \textbf{es} & \textbf{fr} & \textbf{it} & \textbf{ja} & \textbf{ru} & \textbf{tr} & \textbf{zh} \\
\midrule
wmt22\_cometkiwi\_qe & \cellcolor[RGB]{237,94,60} \color{white} -0.642 & \cellcolor[RGB]{215,49,39} \color{white} -0.796 & \cellcolor[RGB]{253,212,129}  -0.249 & \cellcolor[RGB]{246,124,74} \color{white} -0.548 & \cellcolor[RGB]{217,53,41} \color{white} -0.774 & \cellcolor[RGB]{253,188,109}  -0.339 & \cellcolor[RGB]{244,114,69} \color{white} -0.586 & \cellcolor[RGB]{251,159,90}  -0.441 & \cellcolor[RGB]{254,243,171}  -0.071 \\
\midrule
wmt23\_cometkiwi\_xl & \cellcolor[RGB]{252,172,96}  -0.403 & \cellcolor[RGB]{225,70,49} \color{white} -0.720 & \cellcolor[RGB]{254,254,189}  -0.005 & \cellcolor[RGB]{245,116,70} \color{white} -0.572 & \cellcolor[RGB]{238,97,61} \color{white} -0.636 & \cellcolor[RGB]{247,131,77} \color{white} -0.529 & \cellcolor[RGB]{250,152,86}  -0.467 & \cellcolor[RGB]{253,180,102}  -0.374 & \cellcolor[RGB]{207,234,132}  0.240 \\
wmt23\_cometkiwi\_xxl & \cellcolor[RGB]{254,225,141}  -0.190 & \cellcolor[RGB]{253,176,99}  -0.384 & \cellcolor[RGB]{171,219,109}  0.377 & \cellcolor[RGB]{253,204,122}  -0.277 & \cellcolor[RGB]{254,226,143}  -0.180 & \cellcolor[RGB]{228,244,155}  0.133 & \cellcolor[RGB]{193,228,123}  0.297 & \cellcolor[RGB]{129,201,102}  0.513 & \cellcolor[RGB]{107,191,99}  0.579 \\
\midrule
xcomet\_xl & \cellcolor[RGB]{249,147,84}  -0.482 & \cellcolor[RGB]{244,111,68} \color{white} -0.589 & \cellcolor[RGB]{254,254,189}  0.006 & \cellcolor[RGB]{250,154,88}  -0.460 & \cellcolor[RGB]{237,94,60} \color{white} -0.647 & \cellcolor[RGB]{248,144,83}  -0.491 & \cellcolor[RGB]{249,147,84}  -0.477 & \cellcolor[RGB]{253,180,102}  -0.373 & \cellcolor[RGB]{228,244,155}  0.136 \\
xcomet\_xxl & \cellcolor[RGB]{199,231,127}  0.266 & \cellcolor[RGB]{254,224,139}  -0.201 & \cellcolor[RGB]{117,195,100}  0.549 & \cellcolor[RGB]{254,254,189}  -0.006 & \cellcolor[RGB]{242,249,173}  0.065 & \cellcolor[RGB]{209,235,133}  0.230 & \cellcolor[RGB]{107,191,99}  0.584 & \cellcolor[RGB]{87,181,95} \color{white} 0.637 & \cellcolor[RGB]{60,168,88} \color{white} 0.710 \\
\midrule
metricx\_23\_qe\_large & \cellcolor[RGB]{247,131,77} \color{white} -0.524 & \cellcolor[RGB]{217,53,41} \color{white} -0.776 & \cellcolor[RGB]{250,154,88}  -0.457 & \cellcolor[RGB]{242,106,65} \color{white} -0.606 & \cellcolor[RGB]{188,22,38} \color{white} -0.902 & \cellcolor[RGB]{225,70,49} \color{white} -0.722 & \cellcolor[RGB]{244,111,68} \color{white} -0.587 & \cellcolor[RGB]{247,131,77} \color{white} -0.526 & \cellcolor[RGB]{230,80,53} \color{white} -0.694 \\
metricx\_23\_qe\_xl & \cellcolor[RGB]{253,196,115}  -0.312 & \cellcolor[RGB]{250,157,89}  -0.446 & \cellcolor[RGB]{253,182,104}  -0.366 & \cellcolor[RGB]{231,82,54} \color{white} -0.684 & \cellcolor[RGB]{245,121,72} \color{white} -0.559 & \cellcolor[RGB]{239,99,62} \color{white} -0.628 & \cellcolor[RGB]{249,149,85}  -0.474 & \cellcolor[RGB]{254,243,171}  -0.078 & \cellcolor[RGB]{254,236,159}  -0.124 \\
metricx\_23\_qe\_xxl & \cellcolor[RGB]{254,254,189}  -0.001 & \cellcolor[RGB]{253,190,110}  -0.332 & \cellcolor[RGB]{254,248,179}  -0.046 & \cellcolor[RGB]{253,186,107}  -0.346 & \cellcolor[RGB]{251,162,91}  -0.433 & \cellcolor[RGB]{253,210,127}  -0.253 & \cellcolor[RGB]{253,218,134}  -0.219 & \cellcolor[RGB]{221,240,145}  0.178 & \cellcolor[RGB]{199,231,127}  0.269 \\
\midrule
metricx\_24\_hybrid\_large & \cellcolor[RGB]{254,226,143}  -0.186 & \cellcolor[RGB]{249,147,84}  -0.481 & \cellcolor[RGB]{254,239,165}  -0.100 & \cellcolor[RGB]{253,200,119}  -0.292 & \cellcolor[RGB]{228,75,51} \color{white} -0.703 & \cellcolor[RGB]{254,236,159}  -0.122 & \cellcolor[RGB]{243,250,175}  0.059 & \cellcolor[RGB]{246,251,179}  0.044 & \cellcolor[RGB]{203,232,129}  0.256 \\
metricx\_24\_hybrid\_xl & \cellcolor[RGB]{228,244,155}  0.133 & \cellcolor[RGB]{164,216,105}  0.403 & \cellcolor[RGB]{175,220,111}  0.362 & \cellcolor[RGB]{240,249,171}  0.071 & \cellcolor[RGB]{209,235,133}  0.231 & \cellcolor[RGB]{217,239,139}  0.198 & \cellcolor[RGB]{162,215,105}  0.413 & \cellcolor[RGB]{157,213,105}  0.427 & \cellcolor[RGB]{84,180,94} \color{white} 0.647 \\
metricx\_24\_hybrid\_xxl & \cellcolor[RGB]{177,221,113}  0.355 & \cellcolor[RGB]{90,183,96} \color{white} 0.629 & \cellcolor[RGB]{96,186,97} \color{white} 0.615 & \cellcolor[RGB]{169,218,107}  0.387 & \cellcolor[RGB]{129,201,102}  0.509 & \cellcolor[RGB]{112,193,100}  0.564 & \cellcolor[RGB]{5,113,59} \color{white} 0.958 & \cellcolor[RGB]{78,177,93} \color{white} 0.660 & \cellcolor[RGB]{22,145,76} \color{white} 0.822 \\
\midrule
gemba\_qwen3\_0.6b\_noThink & \cellcolor{black} \textcolor{white}{\textbf{N/A}} & \cellcolor{black} \textcolor{white}{\textbf{N/A}} & \cellcolor{black} \textcolor{white}{\textbf{N/A}} & \cellcolor[RGB]{254,242,169}  -0.079 & \cellcolor[RGB]{228,244,155}  0.136 & \cellcolor[RGB]{254,224,139}  -0.198 & \cellcolor[RGB]{254,247,177}  -0.047 & \cellcolor[RGB]{253,214,130}  -0.236 & \cellcolor[RGB]{247,134,78} \color{white} -0.523 \\
gemba\_qwen3\_0.6b & \cellcolor[RGB]{254,243,171}  -0.071 & \cellcolor[RGB]{253,218,134}  -0.226 & \cellcolor[RGB]{254,243,171}  -0.072 & \cellcolor[RGB]{254,233,155}  -0.140 & \cellcolor[RGB]{253,194,114}  -0.313 & \cellcolor[RGB]{254,231,151}  -0.153 & \cellcolor[RGB]{254,226,143}  -0.185 & \cellcolor[RGB]{254,238,163}  -0.106 & \cellcolor[RGB]{254,253,187}  -0.010 \\
gemba\_qwen3\_1.7b\_noThink & \cellcolor[RGB]{232,85,56} \color{white} -0.675 & \cellcolor[RGB]{224,68,48} \color{white} -0.732 & \cellcolor[RGB]{248,252,181}  0.035 & \cellcolor[RGB]{253,202,120}  -0.283 & \cellcolor[RGB]{253,200,119}  -0.295 & \cellcolor[RGB]{254,234,157}  -0.127 & \cellcolor[RGB]{253,204,122}  -0.280 & \cellcolor[RGB]{247,134,78} \color{white} -0.521 & \cellcolor[RGB]{171,219,109}  0.377 \\
gemba\_qwen3\_1.7b & \cellcolor[RGB]{246,129,76} \color{white} -0.538 & \cellcolor[RGB]{233,87,57} \color{white} -0.671 & \cellcolor[RGB]{254,251,185}  -0.018 & \cellcolor[RGB]{253,206,124}  -0.266 & \cellcolor[RGB]{254,230,149}  -0.160 & \cellcolor[RGB]{254,238,163}  -0.106 & \cellcolor[RGB]{253,220,135}  -0.212 & \cellcolor[RGB]{253,194,114}  -0.313 & \cellcolor[RGB]{109,192,99}  0.573 \\
gemba\_qwen3\_8b\_noThink & \cellcolor[RGB]{253,206,124}  -0.273 & \cellcolor[RGB]{222,63,46} \color{white} -0.743 & \cellcolor[RGB]{199,231,127}  0.272 & \cellcolor[RGB]{159,214,105}  0.416 & \cellcolor[RGB]{219,240,143}  0.185 & \cellcolor[RGB]{147,208,103}  0.458 & \cellcolor[RGB]{134,203,102}  0.497 & \cellcolor[RGB]{179,222,114}  0.344 & \cellcolor[RGB]{45,161,84} \color{white} 0.749 \\
gemba\_qwen3\_8b & \cellcolor[RGB]{253,182,104}  -0.361 & \cellcolor[RGB]{221,61,45} \color{white} -0.757 & \cellcolor[RGB]{193,228,123}  0.292 & \cellcolor[RGB]{181,223,115}  0.337 & \cellcolor[RGB]{217,239,139}  0.201 & \cellcolor[RGB]{175,220,111}  0.365 & \cellcolor[RGB]{152,210,104}  0.439 & \cellcolor[RGB]{245,250,177}  0.051 & \cellcolor[RGB]{42,159,84} \color{white} 0.750 \\
gemba\_qwen3\_14b\_noThink & \cellcolor[RGB]{248,252,181}  0.033 & \cellcolor[RGB]{244,109,67} \color{white} -0.594 & \cellcolor[RGB]{193,228,123}  0.295 & \cellcolor[RGB]{122,197,101}  0.532 & \cellcolor[RGB]{207,234,132}  0.237 & \cellcolor[RGB]{66,171,90} \color{white} 0.694 & \cellcolor[RGB]{114,194,100}  0.558 & \cellcolor[RGB]{179,222,114}  0.347 & \cellcolor[RGB]{17,136,71} \color{white} 0.867 \\
gemba\_qwen3\_14b & \cellcolor[RGB]{254,254,189}  -0.006 & \cellcolor[RGB]{248,144,83}  -0.485 & \cellcolor[RGB]{195,229,124}  0.285 & \cellcolor[RGB]{185,225,118}  0.322 & \cellcolor[RGB]{187,226,119}  0.313 & \cellcolor[RGB]{96,186,97} \color{white} 0.616 & \cellcolor[RGB]{157,213,105}  0.426 & \cellcolor[RGB]{236,247,165}  0.099 & \cellcolor[RGB]{42,159,84} \color{white} 0.757 \\
gemba\_qwen3\_32b\_noThink & \cellcolor[RGB]{254,236,159}  -0.125 & \cellcolor[RGB]{222,63,46} \color{white} -0.746 & \cellcolor[RGB]{175,220,111}  0.365 & \cellcolor[RGB]{154,212,104}  0.431 & \cellcolor[RGB]{199,231,127}  0.268 & \cellcolor[RGB]{93,184,96} \color{white} 0.619 & \cellcolor[RGB]{119,196,100}  0.540 & \cellcolor[RGB]{254,243,171}  -0.076 & \cellcolor[RGB]{13,128,67} \color{white} 0.892 \\
gemba\_qwen3\_32b & \cellcolor[RGB]{243,250,175}  0.056 & \cellcolor[RGB]{253,180,102}  -0.374 & \cellcolor[RGB]{173,220,110}  0.369 & \cellcolor[RGB]{164,216,105}  0.399 & \cellcolor[RGB]{195,229,124}  0.288 & \cellcolor[RGB]{84,180,94} \color{white} 0.645 & \cellcolor[RGB]{139,205,103}  0.482 & \cellcolor[RGB]{254,253,187}  -0.010 & \cellcolor[RGB]{18,137,72} \color{white} 0.856 \\
\midrule
gemba\_gemma4\_e2b\_noThink & \cellcolor[RGB]{240,249,171}  0.072 & \cellcolor[RGB]{228,75,51} \color{white} -0.708 & \cellcolor[RGB]{104,190,99}  0.593 & \cellcolor[RGB]{84,180,94} \color{white} 0.644 & \cellcolor[RGB]{240,249,171}  0.077 & \cellcolor[RGB]{69,173,90} \color{white} 0.684 & \cellcolor[RGB]{84,180,94} \color{white} 0.641 & \cellcolor[RGB]{112,193,100}  0.570 & \cellcolor[RGB]{12,126,66} \color{white} 0.899 \\
gemba\_gemma4\_e2b & \cellcolor[RGB]{253,202,120}  -0.281 & \cellcolor[RGB]{223,65,47} \color{white} -0.737 & \cellcolor[RGB]{134,203,102}  0.494 & \cellcolor[RGB]{139,205,103}  0.482 & \cellcolor[RGB]{254,237,161}  -0.112 & \cellcolor[RGB]{132,202,102}  0.506 & \cellcolor[RGB]{127,199,101}  0.517 & \cellcolor[RGB]{201,232,128}  0.265 & \cellcolor[RGB]{36,157,82} \color{white} 0.769 \\
gemba\_gemma4\_e4b\_noThink & \cellcolor[RGB]{199,231,127}  0.270 & \cellcolor[RGB]{245,121,72} \color{white} -0.555 & \cellcolor[RGB]{72,174,91} \color{white} 0.679 & \cellcolor[RGB]{66,171,90} \color{white} 0.694 & \cellcolor[RGB]{171,219,109}  0.377 & \cellcolor[RGB]{22,145,76} \color{white} 0.826 & \cellcolor[RGB]{18,137,72} \color{white} 0.859 & \cellcolor[RGB]{104,190,99}  0.587 & \cellcolor[RGB]{15,132,69} \color{white} 0.881 \\
gemba\_gemma4\_e4b & \cellcolor[RGB]{215,238,137}  0.208 & \cellcolor[RGB]{251,162,91}  -0.432 & \cellcolor[RGB]{66,171,90} \color{white} 0.690 & \cellcolor[RGB]{84,180,94} \color{white} 0.648 & \cellcolor[RGB]{181,223,115}  0.343 & \cellcolor[RGB]{63,170,89} \color{white} 0.698 & \cellcolor[RGB]{33,155,81} \color{white} 0.779 & \cellcolor[RGB]{173,220,110}  0.373 & \cellcolor[RGB]{14,130,68} \color{white} 0.890 \\
gemba\_gemma4\_31b\_noThink & \cellcolor[RGB]{8,119,62} \color{white} 0.937 & \cellcolor[RGB]{66,171,90} \color{white} 0.695 & \cellcolor[RGB]{11,124,65} \color{white} 0.911 & \cellcolor[RGB]{7,117,61} \color{white} 0.941 & \cellcolor[RGB]{14,130,68} \color{white} 0.884 & \cellcolor[RGB]{1,105,55} \color{white} 0.988 & \cellcolor[RGB]{4,111,58} \color{white} 0.968 & \cellcolor[RGB]{0,104,55} \color{white} 0.999 & \cellcolor[RGB]{3,109,57} \color{white} 0.974 \\
gemba\_gemma4\_31b & \cellcolor[RGB]{6,115,60} \color{white} 0.950 & \cellcolor[RGB]{57,167,87} \color{white} 0.715 & \cellcolor[RGB]{12,126,66} \color{white} 0.905 & \cellcolor[RGB]{6,115,60} \color{white} 0.947 & \cellcolor[RGB]{24,149,78} \color{white} 0.807 & \cellcolor[RGB]{0,104,55} \color{white} 0.993 & \cellcolor[RGB]{1,105,55} \color{white} 0.985 & \cellcolor[RGB]{0,104,55} \color{white} 0.999 & \cellcolor[RGB]{3,109,57} \color{white} 0.970 \\
\midrule
gemba\_olmo3\_7b & \cellcolor[RGB]{253,180,102}  -0.374 & \cellcolor[RGB]{245,116,70} \color{white} -0.575 & \cellcolor[RGB]{254,236,159}  -0.120 & \cellcolor[RGB]{253,216,132}  -0.228 & \cellcolor[RGB]{253,182,104}  -0.361 & \cellcolor[RGB]{253,194,114}  -0.315 & \cellcolor[RGB]{253,212,129}  -0.249 & \cellcolor[RGB]{253,216,132}  -0.228 & \cellcolor[RGB]{224,242,149}  0.162 \\
gemba\_olmo3\_32b & \cellcolor[RGB]{252,172,96}  -0.400 & \cellcolor[RGB]{248,142,82}  -0.493 & \cellcolor[RGB]{233,245,161}  0.111 & \cellcolor[RGB]{254,232,153}  -0.145 & \cellcolor[RGB]{253,204,122}  -0.275 & \cellcolor[RGB]{254,224,139}  -0.202 & \cellcolor[RGB]{254,231,151}  -0.154 & \cellcolor[RGB]{246,251,179}  0.046 & \cellcolor[RGB]{84,180,94} \color{white} 0.642 \\
\bottomrule
\end{tabular}%
}
\caption{Mean Kendall's $\tau$ rank correlation coefficient demonstrating adherence to the idiomaticity gradient. Values closer to $1.0$ indicate a perfect monotonic preference for idiomaticity over literality. N/A values denote instances where a model collapsed into outputting a single, constant score for all permutations within a segment, rendering rank correlation mathematically undefined.}
\label{tab:kendalls_tau}
\end{table*}

As previously stated, judgments of literality and idiomaticity exist on a continuous scale rather than as a strictly binary dichotomy. To evaluate the systems' sensitivity to more granular degrees of idiomaticity, we analyzed their performance on the graded $L=3$ dataset partition we compiled, which features three-sentence segments representing various degrees of literality/idiomaticity (100\%, 66\%, 33\%, and 0\% literal permutations; see Table~\ref{tab:idiomaticity_gradient}). For each three-sentence instance, we calculated Kendall's $\tau_b$ rank correlation coefficient \citep{kendall-tau-1938} to measure how closely a model's predicted scores aligned with the ideal 4-step idiomaticity gradient the segments represent. Kendall's $\tau$ evaluates the ordinal association between two measured quantities; in the context of this experiment, a $\tau$ of $1.0$ indicates a perfect monotonic preference for idiomaticity (correctly ranking all four steps of the gradient in the right order), a $\tau$ near $0.0$ indicates random scoring, and a negative $\tau$ denotes an active preference for literal translations. The mean segment-level $\tau$ scores are displayed in Table~\ref{tab:kendalls_tau} and broadly corroborate the binary macro-preference trends: older ``conventional'' QE metrics and smaller LLM checkpoints exhibit strong negative correlations, heavily penalizing idiomaticity. Crucially, however, evaluating rank order across intermediate steps provides the resolution needed to separate the top-performing systems. While metricx\_24\_hybrid\_xxl and gemba\_gemma4\_31b appeared roughly on par when evaluating the binary extremes of full literality and idiomaticity, this Kendall's $\tau$ analysis reveals the performance boundaries of the ``conventional'' QE metric. Metricx\_24\_hybrid\_xxl struggles to reliably rank the intermediate 33\% and 66\% sentence mixtures, resulting in only moderate correlations on languages like Arabic ($\tau = 0.355$) and French ($\tau = 0.387$). In stark contrast, gemba\_gemma4\_31b maintains near-perfect monotonic tracking across the board, achieving $\tau$ scores above $0.900$ for the vast majority of languages. This suggests that while large conventional systems reliably capture extreme contrasts, the 31B LLM is substantially better equipped to track subtle ``real-world'' gradations of idiomaticity and literality.

\section{Related Work}
The state of the art in QE research was shaped through successive editions of WMT shared tasks beginning in 2012 \citep{callison-burch-etal-2012-findings,bojar-etal-2013-findings,bojar-etal-2014-findings,bojar-etal-2015-findings,bojar-etal-2016-findings,bojar-etal-2017-findings,specia-etal-2018-findings,fonseca-etal-2019-findings,specia-etal-2020-findings-wmt,specia-etal-2021-findings,zerva-etal-2022-findings,blain-etal-2023-findings,zerva-etal-2024-findings,lavie-etal-2025-findings}, though QE as a research program had substantial precedents under the banner of ``confidence estimation'' \citep[\textit{inter alia}][]{gandrabur-foster-2003-confidence,quirk-2004-training,blatz-etal-2004-confidence,ueffing-ney-2005-word,specia-etal-2009-estimating,xiong-etal-2010-error,bach-etal-2011-goodness}. \citet{Zhao-survey-2024} offer a comprehensive survey of QE research, grouping approaches into three categories that also reflect the historical evolution of the task: handcrafted-feature-based systems \citep[\textit{inter alia}]{specia-etal-2013-quest,specia-etal-2015-multi}, deep-learning-based systems---a category subdivided into ``classic deep learning methods'' \citep[\textit{inter alia}]{kreutzer-etal-2015-quality,kim-lee-2016-recurrent,kim-etal-2017-predictor,ive-etal-2018-deepquest,martins-etal-2017-pushing} and ``methods incorporating pre-trained language models'', the subcategory to which COMET and MetricX systems belong---and LLM-based approaches, to which the GEMBA framework belongs.

Recent studies have begun probing the specific failure modes and biases of QE systems. Work by \citet{zaranis-etal-2025-watching}, which is closest to our own in its methodology, audited the gender biases of QE systems, showing that they systematically favor masculine-inflected translations when gender is ambiguous in source segments. Contemporaneously, \citet{tian-etal-2026-beyond} investigated the broader challenges reference-based and reference-free metrics face when evaluating non-literal translations (spanning social media slang and puns, culture-specific \textit{realia}, and the literary domain) for the English-Chinese language pair, demonstrating that traditional QE metrics correlate poorly with human preferences in non-literal translation contexts, whereas LLM-as-a-judge approaches align most closely. To the best of our knowledge, ours is the first study of QE focused strictly on idiom translation, though the topic of idiom translation itself has received ample treatment in MT research \citep[\textit{inter alia}]{wehrli-1998-translating-idioms,anastasiou2010idiom,salton-etal-2014-evaluation,salton-etal-2014-empirical,martins-etal-2017-pushing,santing-etal-2022-food,zaitova-etal-2025-walk,wang-etal-2025-proverbs,gasparovic-etal-2025-assessing,thanh-nguyen-kordjamshidi-2025-dont,donthi-etal-2025-improving,arslan-etal-2025-using,rezaeimanesh-etal-2025-large,jiang-etal-2026-derea,montesinos-etal-2026-improving}.

\citet{flor-etal-2025-survey} present an extensive inventory of idiom corpora,\footnote{Also maintained as an updated online repository: \url{https://github.com/maafiah/IdiomsResearch}} distinguishing between datasets compiled for the purpose of psycholinguistic and second language acquisition research on the one hand \citep[19 corpora; \textit{inter alia,}][]{cronk1993idioms,libben2008multidetermined,caillies2009descriptions,Morid_Sabourin_2024}, and natural language processing research on the other hand (34 corpora). The latter category comprises a wide range of datasets, and includes a few parallel corpora, with idioms used in context and translated idiomatically into a target language \citep{fadaee-etal-2018-examining,stap-etal-2024-fine}, idioms in their dictionary citation form with idiomatic translations in target languages \citep{tang2022petciparallelenglishtranslation,agrawal-etal-2018-beating,moussallem-etal-2018-lidioms,li-idiomkb-2024}, as well as monolingual ``parallel'' corpora consisting in idioms paired with an explicit paraphrase in the same language \citep{zhou-etal-2021-pie,liu-hwa-2016-phrasal,pershina-etal-2015-idiom}, with \citet{tayyar-madabushi-etal-2021-astitchinlanguagemodels-dataset} additionally providing literal paraphrases of idioms in the same language for a subset of their corpus.\footnote{Another dataset offering such annotations was recently compiled by \citet{almheiri-etal-2026-multilingual}.} The dataset closest to our own is the concurrently compiled G-IdiomAlign benchmark \citep{ye-etal-2026-g}, which pairs cross-lingual idioms automatically retrieved from Wiktionary in their dictionary form and notably includes literal translations to evaluate LLM bias toward literal transfer. Our own work complements it in at least one critical way, by featuring idioms embedded within full sentences, as approaches relying on absolute dictionary citation forms naturally tend to abstract away the discourse often required to disambiguate literal from figurative usage. To the best of our knowledge, ours is the first dataset to align idiom-containing source sentences with idiomatic and literal translations across a graded continuum.

\section{Conclusion}

We introduced a novel parallel idiom dataset of paired literal and idiomatic translations across nine languages, and used it to study the performance of QE systems in idiom translation. Our results show that while smaller models and older QE metrics exhibit a marked literality bias, scaling up model parameters reliably improves idiomatic sensitivity. Our findings also reveal that while the largest conventional QE metrics can successfully evaluate binary extremes of literality/idiomaticity, LLMs in the GEMBA regime are uniquely capable of accurately ranking more granular, intermediate gradations of these textual properties.

\section*{Limitations}
\paragraph{Dataset coverage and construction} English is our only source language. Our results may therefore overestimate QE performance on rarer, more culture-specific idioms, and they may not transfer to other source languages. Our nine target languages are also comparatively well resourced. The triplets were curated by the authors and native-speaker colleagues drawing on established lexicographical resources but without a formal validation protocol or inter-annotator agreement measure. Finally, the concatenated segments are generally topically incoherent rather than naturally occurring discourse, which might restrict this dataset's usefulness for other downstream tasks.

\paragraph{Model and prompt coverage} We only evaluated open-weight systems (see Ethics Statement). Proprietary LLMs are therefore absent from our study in spite of their wide use as QE judges. Each LLM was queried with a single zero-shot DA-style prompt under greedy decoding. Other regimes, such as MQM-style error annotation, few-shot prompting or prompts that mention idiomaticity explicitly, may produce different behaviors. Finally, our observations about scale rest on few checkpoints per family, and size might be confounded with other factors such as distillation.

\section*{Ethics Statement}

We acknowledge the environmental impact of large-scale LLM inference, particularly with the reasoning modes enabled. We had initially planned a further comparison between literality/idiomaticity judgments and QE sensitivity to synthetic error patterns (word deletion and permutation, character deletion and permutation) in idiom-bearing sentences at several degrees of corruption, but chose to abandon this analysis as its compute cost and environmental impact did not appear justified in light of the limited insight it was expected to offer. Our work consumed allocations on a cluster that operates on largely decarbonized electricity, and we release all model outputs under CC licenses for re-use by the community (see below).

The dataset was constructed by the authors and consenting colleagues on a voluntary basis, with no compensation involved and no personal data collected. We screened the triplets for offensive content, but idioms can encode cultural or ethnic stereotypes, and some expressions in the dataset may carry such connotations.

We used all models in accordance with their licenses and intended research use, limiting our experiments to open-weight models with permissive licenses to this specific end. We release the dataset under a CC BY-SA 4.0 license. We also release all LLM completions, including reasoning traces, and all QE metric scores under CC BY-NC-SA 4.0,\footnote{In line with the least permissive license of those checkpoints.} so that our results can be reproduced and reused by the community. The dataset is intended as an evaluation resource and will be published as a password-protected archive to prevent its inclusion in web crawls; we discourage---but do not prohibit---its use as training data for QE and MT systems.

QE systems are increasingly used in professional translation workflows. Our finding that many of them prefer literal over idiomatic renderings implies that, in such settings, skillful human translations could be scored below poor literal machine output. We encourage practitioners to keep human oversight in place wherever QE scores inform decisions about translated content or about translators’ work, especially for text domains rich in figurative language.

\section*{Acknowledgments}

This work was funded by the French \textit{Agence Nationale de la Recherche} (ANR) under the project TraLaLaM (``ANR-23-IAS1-0006''), as well as by Inria under the ``\textit{Défi}''-type project COLaF.  It was also partly funded by Benoît Sagot’s chair in the PRAIRIE-PSAI institute, also funded by the French national agency ANR as part of the “France 2030” strategy under the reference ANR-23-IACL-0008.

This work was also granted access to the HPC resources of IDRIS under allocations 2025-AD011015117R2, 2025-AD011012254R5 and 2025-AD011017228 made by GENCI. The authors acknowledge the use of LLMs as coding assistants during the implementation of this work.

\bibliography{custom}

\appendix
\section{GEMBA-Style Prompt Template}
\label{app:prompt}

The following are unpopulated examples of the system and user GEMBA-style prompts used to obtain direct assesment QE scores from various models of the Qwen-3, Olmo-3 and Gemma-4 model families. The placeholders \texttt{\{tgt\_lang\}}, \texttt{\{src\}}, and \texttt{\{hyp\}} stand for ``target language'' (one of the nine in our custom dataset), ``source segment'' (an English source segment from our dataset) and ``translation hypothesis'' (one of the variously literal or idiomatic translations in our dataset), respectively. Two prompt templates were used to either encourage or prohibit reasoning/chain of thought; the LLMs' reasoning mode were further triggered through the modalities prescribed by their developers (e.g., through the injection of a ``\texttt{<|think|>}'' token at the beginning of the Gemma-4 system prompts).

\vspace{0.5cm}

\begin{promptbox}[title=System Prompt]

You are an expert translator and translation evaluator.

\end{promptbox}

\vspace{0.5cm}

\begin{promptbox}[title=User Prompt Template (Reasoning On)]

Score the following translation from English to \texttt{\{tgt\_lang\}} on a continuous scale from 0 to 100, where a score of zero means ``no meaning preserved'' and a score of one hundred means ``perfect meaning and grammar.'' \textbf{You may reflect on the task at hand prior to stating your score.} You must output your final score strictly between XML tags thusly: \texttt{<score>your score here</score>}, with only the score digits and no extraneous text between the tags. \\

English source: ``\texttt{\{src\}}''\\
\texttt{\{tgt\_lang\}} translation: ``\texttt{\{hyp\}}''

\end{promptbox}

\vspace{0.5cm}

\begin{promptbox}[title=User Prompt Template (Reasoning Off)]

Score the following translation from English to \texttt{\{tgt\_lang\}} on a continuous scale from 0 to 100, where a score of zero means ``no meaning preserved'' and a score of one hundred means ``perfect meaning and grammar.'' 
 You must output your final score strictly between XML tags thusly: \texttt{<score>your score here</score>}. \textbf{Provide ONLY the score and the tags, with no additional text, explanation, or reflection.}\\

English source: ``\texttt{\{src\}}''\\
\texttt{\{tgt\_lang\}} translation: ``\texttt{\{hyp\}}''

\end{promptbox}

\section{Bootstrap Significance Test}
\label{app:bootstrap}

To compare the performance of all evaluated QE systems, we conducted a stratified bootstrap resampling analysis ($N=1000$ iterations), computing the statistical significance of the differences in macro-preference ratios between every pair of systems while ensuring that each language pair contributes equally to the global evaluation. For quality metrics (COMET, GEMBA scale), this is calculated as the mean score of full literal translations divided by the mean score of full idiomatic translations. For error-based metrics (MetricX), the ratio is inverted to maintain reporting consistency. For each bootstrap iteration, the global macro-preference ratio was computed for every system. We then calculated the pairwise differences ($\Delta$) by subtracting the column system's ratio from the row system's ratio. Two-tailed $p$-values were derived by measuring the proportion of bootstrap iterations in which the calculated $\Delta$ crossed zero. Differences that achieved statistical significance ($p < 0.05$) are explicitly annotated with an asterisk (*) in the resulting heatmap. The resulting matrix provides a comprehensive system-wide comparison of literality bias.  The numerical value inside each cell represents the mean $\Delta$ Macro-Preference Ratio (Row - Column).  The diverging color palette (Red/Blue) visually encodes this difference. Red cells indicate a positive delta, meaning the row system exhibits a stronger literality bias than the column system. Conversely, blue cells indicate a negative delta, meaning the row system rates idiomatic translations higher than literal ones by a larger margin than the column system.  

\begin{figure*}
    \centering
    \includegraphics[width=1\linewidth]{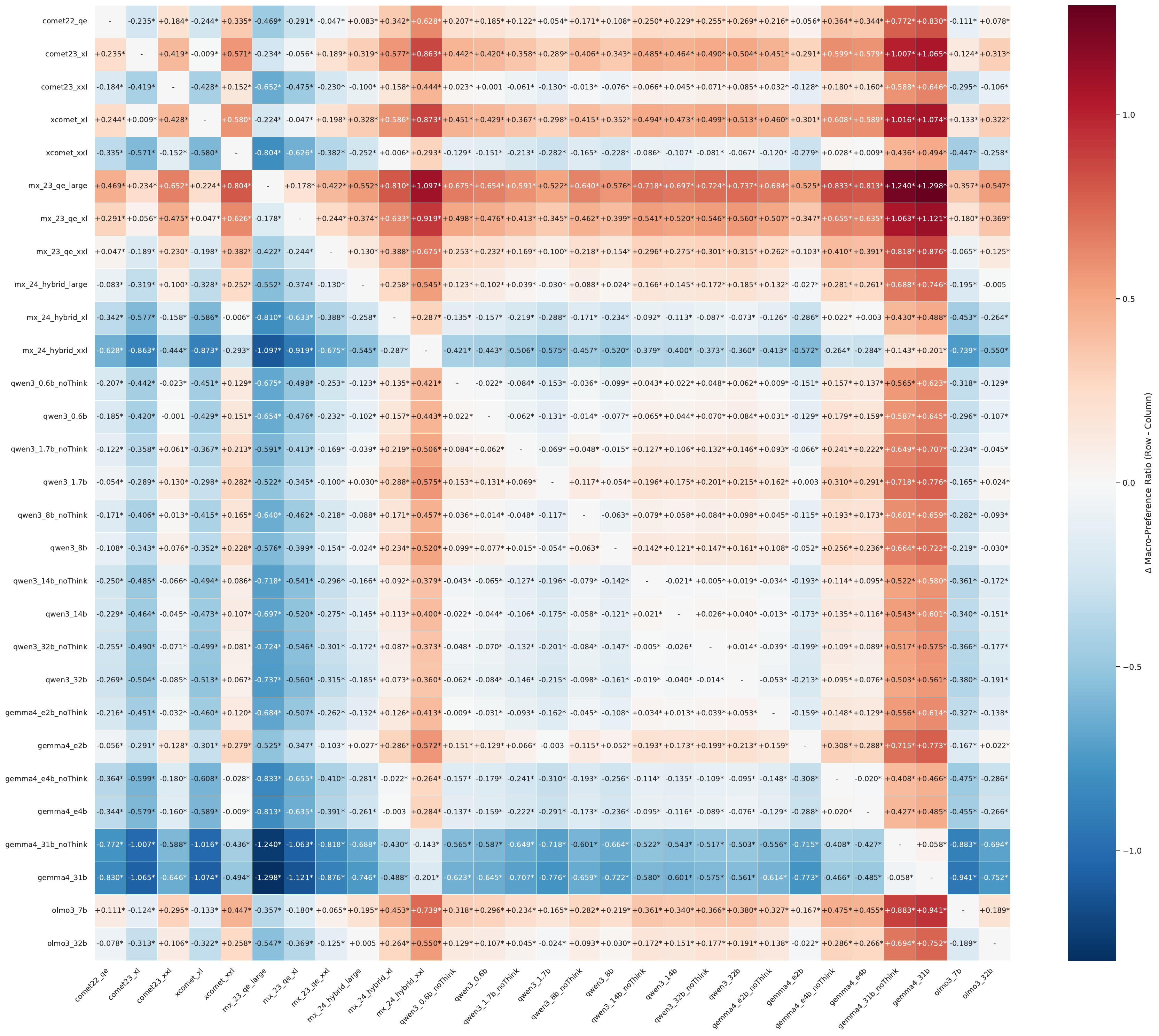}
    \caption{System-Wide Comparison of Preference Ratio Deltas; blue means row rates idiomatic higher than literal by a larger margin than column, and red the opposite; * denotes p < 0.05 via Stratified Bootstrap).}
    \label{fig:heatmap_bootstrap}
\end{figure*}

\end{document}